\documentclass[10pt,twocolumn,letterpaper]{article}

\usepackage{cvpr}
\usepackage{times}
\usepackage{epsfig}
\usepackage{graphicx}
\usepackage{amsmath}
\usepackage{amssymb}
\usepackage{subcaption}

\usepackage[pagebackref=true,breaklinks=true,letterpaper=true,colorlinks,bookmarks=false]{hyperref}

\cvprfinalcopy 

\def\cvprPaperID{****} 

\ifcvprfinal\fi
\begin{document}

\title{SuperOCR: A Conversion from Optical Character Recognition\\to Image Captioning}

\author{Baohua Sun, Michael Lin, Hao Sha, Lin Yang\\
Gyrfalcon Technology Inc.\\
1900 McCarthy Blvd Suite 208, Milpitas, CA, 95035\\
{\tt\small baohua.sun@gyrfalcontech.com}
}

\maketitle

\begin{abstract}
   Optical Character Recognition (OCR) has many real world applications. The existing methods normally detect where the characters are, and then recognize the character for each detected location. Thus the accuracy of characters recognition is impacted by the performance of characters detection. In this paper, we propose a method for recognizing characters without detecting the location of each character. This is done by converting the OCR task into an image captioning task. One advantage of the proposed method is that the labeled bounding boxes for the characters are not needed during training. The experimental results show the proposed method outperforms the existing methods on both the license plate recognition and the watermeter character recognition tasks. The proposed method is also deployed into a low-power (300mW) CNN accelerator chip connected to a Raspberry Pi 3 for on-device applications.

\end{abstract}

\section{Introduction}

Optical Characters Recognition (OCR) has many real-world applications. Depending on the application scenarios, there are mainly two types of OCR tasks. The first type of OCR task is the separated single character recognition, such as MNIST~\cite{lecun1998mnist} and CASIA handwritten Chinese character recognition~\cite{liu2011casia}, which are generally treated as a classification problem; the second OCR task is to recognize a string of characters at random locations with various backgrounds in the image, such as Street View House Number recognition (SVHN)~\cite{netzer2011reading} and CASIA handwritten Chinese text recognition~\cite{liu2011casia}. Specially, the number of characters are fixed in some application scenarios, such as license plate recognition~\cite{xu2018towards} and watermeter characters recognition~\cite{yang2019fully}. In this paper, we focus on the second type of OCR task with fixed number of characters. 

Current methods~\cite{laroca2019efficient,laroca2018robust,silva2017real,selmi2017deep,rizvi2017deep,masood2017license} for the second type of OCR task normally detect each character first as an intermediate result, and then recognize the character in each detected location. For real-world applications, such as license plate recognition and watermeter numbers recognition, people are only interested in the characters reading, instead of the locations of the characters. Recent research~\cite{xu2018towards,li2018toward} also study the end-to-end method to simultaneously detect and recognize the characters. However, besides the characters recognition module, the characters detection module is still designed within the DNN model. 

There are problems with the current methods for both the training phase and the inference phase. During the training phase, the labeled location information, which is usually given in the format of bounding boxes, are required in order to train the model. However, for the majority of real-world applications, the interested output for the OCR tasks are the readings of the characters in the image, instead of where these characters are located. These location information during inference phase are only used as intermediate result, and will be of no use once the characters recognition is completed. However, due to the design of current methods, the location labels are mandate in order to train the model, which unnecessarrily increase the cost of obtaining the labeled traing data. In fact, labeling the bounding boxes of the characters is much heavier work than only labeling the readings of the characters. In real-world application scenarios, the boundary of the location may also be rotated~\cite{xu2018towards}. Therefore, some extra efforts for labeling the characters location in quadrilateral areas are also needed in addition to the bounding boxes, which adds more costs for labeling. As a result, the existing methods are very expensive to train. For a task only interested in the characters reading, these human labors could have been saved if the algorithm design requires no characters location information for training. 

During the inference phase, the detection module also has to calculate the location of the characters in order for the recognition module to work on, which results in an extra computation power consumption. The existing methods are usually deployed on GPUs which consumes large volumes of power. In spite of recent years development and wide availability of low-power CNN accelerators~\cite{sun2018ultra,sun2018mram}, the existing methods are hard to be fully deployed into the edge devices. This is because the detection module usually has a different network architecture from that of the image recognition module, which also makes the inference difficult to be deployed on a single accelerator chip to accelerate the heterogeneous network. One practical compromise is to partially deploy the DNN network on the accelerator chip, and execute the rest of the network on the host processor, e.g. CPU. However, this is not the most power-efficient solution for OCR on the edge device, since the whole network is not fully accelerated, and the power consumption on the host processor is still very high. The ideal implementation should be the entire OCR DNN network loaded into the low-power accelerator, and the host processor acting only as a controller to send the input image and receive the output characters readings. 

This paper proposes a method named SuperOCR to address the above mentioned problems by eliminating the need of detection module in the OCR network. The key idea is to convert the OCR task into an image captioning task.

The OCR task shares some common features with the image captioning task. Image captioning takes an image as input, and outputs the text describing the contents in the image~\cite{johnson2016densecap,anderson2018bottom,hossain2019comprehensive,karpathy2015deep,lu2017knowing,rennie2017self,vinyals2016show,yao2017boosting,you2016image}. Recent work of the SuperCaptioning method~\cite{sun2019supercaptioning} combines the multi-modal information of image and text together, and uses a CNN model to predict the image caption iteratively until a stopping criteria is met. This method is inspired by the recent success of Super Characters method~\cite{sun2018super} in Natural Language Processing (NLP) tasks, and borrows the concept of two-dimensional embedding to project the text as part of an image. The SuperCaptioning method requires no knowledge of the objects location, and only uses a CNN model for image recognition in each iteration. Similar to the OCR tasks addressed in this paper, the image captioning task also takes an image as input, and outputs a string of characters or words. 

In spite of these similarities, the OCR task and the image captioning task are different in some ways. First, the answer for an OCR task is normally objective and unique. However, this is not neccessary for the image captioning task, which is more subjective and non-exclusive. An image can have multiple correct captions given by different people from different perspectives of the image; Second, the focus of an OCR task could be a small area located in a large background, e.g. recognizing the house numbers in a street view image. While the caption for an image may either pay attention to a small area or describes only the general scenary; Third, some OCR applications have fixed number of characters to be recognized, such as watermeter character recognition and license plate recognition. However, the image captioning task normally has no such constraint. These differences bring challenges when treating the OCR problem as an Image Captioning problem.


The proposed SuperOCR method casts the OCR problem as a constrained image captioning problem, and modifies the SuperCaptioning method to adapt to the differences between these two problems. The contribution of this paper is three-folded. First, to the best of our knowledge, this is the first paper to address the OCR problem by converting it into an image captioning problem; Second, similar to the Super Captioning method which requires no detection module, the SuperOCR method recognizes the characters in random locations without detecting the locations of the characters. This also lowers the cost of building the OCR training set because it saves the labor for the labeled bounding boxes of the characters; Third, the proposed method is also deployed on a Raspberry Pi 3 with a 300mw low-power CNN accelerator chip. Experimental results on car license plate recognition dataset CCPD~\cite{xu2018towards} shows our method obtains an accuracy of 98.7\% on the CCPD-Base dataset, while the best existing method performance is 98.5\%. The significant improved accuracy is observed on more challenging tasks of CCPD-Rotate(95.6\% over 90.8\%), CCPD-Weather(94.7\% over 87.9\%), and CCPD-Tilt(96.0\% over 92.5\%). Another experiment on a watermeter characters recognition dataset~\cite{yang2019fully} also shows our method improves the accuracy from 89.6\% to 90.5\%.

\begin{figure*}[htb]
  \centering
  \centerline{\includegraphics[width=17.5cm]{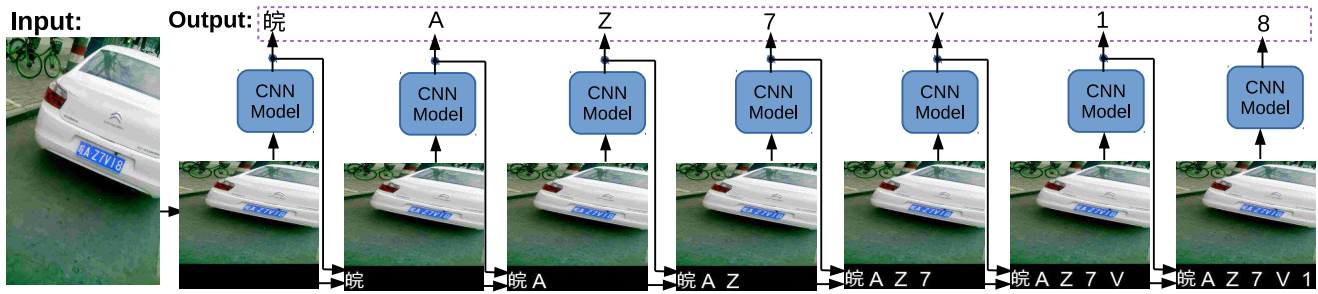}}
\caption{Illustration of the proposed SuperOCR method. The OCR task is converted into an image captioning task, by restricting the image description as the readings of the interested characters found in the image. }
\label{SuperOCR}
\end{figure*}

\section{Proposed Method: SuperOCR}

In this paper, we focus on the second type of OCR task which has a fixed number of characters. The proposed SueprOCR method treats the OCR task as an image captioning problem, and the characters are predicted iteratively by a trained CNN model. This is very similar to the SuperCaptioning~\cite{sun2019supercaptioning} method for generating the image descriptions. 

Fig.~\ref{SuperOCR} illustrates the proposed SuperOCR method during the inference phase, using the Chinese license plate recognition as an example. The input image is first resized to occupy the upper portion of a new image called SuperOCR image, and the lower portion of the SuperOCR image is reserved blank for the characters to be predicted. The SuperOCR image is then fed into a trained CNN model to predict the first character. The first predicted charater will then be drawn onto the bottom left portion to update the SuperOCR image, as shown in Fig.~\ref{SuperOCR}. Again, the updated SuperOCR image will be fed into the same CNN model to predict the next character, and the predicted next character will be drawn on the bottom portion of the SuperOCR image, next to the last predicted chacter. This process repeats for the same number of iterations as the given number of characters.

It can be seen that the proposed SuperOCR method predicts the characters in a sequential manner. During each iteration, the prediction is an image classificaiton problem. The number of classes is determined by the possible values of the characters. This is similar to the vocabulary of words in the image captioning tasks.

In order to train the CNN model in the SuperOCR method, the training dataset containing the SuperOCR images needs to be generated from the labeled OCR dataset. Only the labeled characters readings are utilized during the SuperOCR image generation, while the locations of the characters are not needed. 
     
\subsection{Generate the Labeled SuperOCR Images}
In this paper, we only focus on the OCR task which has a fixed number of characters to be recognized in each input image. For an input image $I$, its ground truth label $Str$ is composed of a string of $N$ characters to represent the characters reading. The proposed SuperOCR method generates $N$ SuperOCR images for the input image $I$. For the $n$\textsuperscript{th} character in $Str$, where $n \in [1,N]$, the $n$\textsuperscript{th} SuperOCR image is generated. This SuperOCR image has the resized $I$ on the upper portion, and also embeds the $(n-1)$ characters before the $n$\textsuperscript{th} character on the lower portion. For the special case of $n=1$, the generated SuperOCR image only has the resized $I$ on the upper portion, but has nothing on the bottom portion. The generated SuperOCR image is labeled as one instance for the class of the $n$\textsuperscript{th} character. 

One important design option during generation is the font size of the characters in the SuperOCR image. Research in Super Characters~\cite{sun2018super} method has shown the impact of characters size on the classification accuracy. It shows that larger font size is beneficial to classification accuracy. For applications like watermeter character recognition and license plate recognition, the number of characters to predict is usually a fixed number. In order to make the best use of the space on the SuperOCR image, the bottom portion of the SupuerOCR image only holds $(N-1)$ characters, instead of $N$. The saved space for the $N$\textsuperscript{th} character will be distributed into the other $(N-1)$ characters, in order to increase the font size of each character for better classification accuracy. For example, a license plate with totally 7 characters will have a design of maximum 6 characters at the lower portion of the SuperOCR image, as shown in Fig.~\ref{SuperOCR}. The 7\textsuperscript{th} SuperOCR image will be labeled as an instance of the class ``8''.
\begin{table*}
\begin{center}
\begin{tabular}{|c|c|c|c|c|c|c|c|c|c|}
\hline
Method   & Resolution & Base & Rotate & Tilt & Weather & DB   & FN   & Challenge\\
  	 &	     & (100k)& (10k)  & (10k)& (10k)   &(20k) & (20k)&(10k)\\
\hline
SSD300~\cite{liu2016ssd}+HC&480x480  &98.3 & 88.4   & 91.5 & 87.3   & 96.6 & {\bf 95.9} & 83.8\%\\
YOLO9000~\cite{redmon2017yolo9000}+HC&480x480  &98.1 & 84.5 & 88.5 & 87.0   & 96.0 & 88.2 & 80.5\%\\
FasterRCNN~\cite{ren2015faster}+HC&480x480  &97.2&82.9 & 87.3 & 85.5   & 94.4 & 90.9 & 76.3\%\\
TE2E~\cite{li2018toward}&480x480  & 97.8&87.9 & 92.1 & 86.8   & 94.8 & 94.5 & 81.2\%\\
RPnet~\cite{xu2018towards}  & 480x480  &98.5 & 90.8   & 92.5 & 87.9   & {\bf 96.9} & 94.3 & {\bf 85.1}\%\\
\hline
SuperOCR &{\bf 310x310}  & {\bf 98.7} & {\bf 95.6}   & {\bf 96.0} & {\bf 94.7}  & 94.6 & 83.6 & 76.9\%\\
\hline
\end{tabular}
\caption{Recognition results on license plate recognition. SuperOCR obtains the highest accuracy for subsets of Base, Rotate, Tilt, and Weather from CCPD datasets, while using a smaller input resolution of 331x331 than the other methods.}
\label{CCPDcomparison}
\end{center}
\end{table*}

\subsection{Train with Generated SuperOCR Images}
The generated SuperOCR training dataset is the same as a typical dataset for the purpose of image classification. The successful CNN network architectures for image classifications could be applied to this dataset through transfer learning. Depending on where the trained model will be deployed, the design choices for model architectures are different. 

If the trained model is to be deployed on general purpose hardwares, such as GPU or CPU, there will be a wide range of CNN model architectures to choose from. One efficient way is to fine-tune pretrained CNN models on ImageNet dataset~\cite{deng2009imagenet}, such as PolyNet~\cite{zhang2017polynet}, EfficientNet~\cite{tan2019efficientnet}, and etc. 

If the target hardware to implement the CNN models are CNN ASIC (Application Specific Integrated Circuits) chips~\cite{sun2018ultra,sun2018mram}, the supported model architures will be limited. In this case, the available model architectures includes VGG~\cite{simonyan2014very}, ResNet~\cite{he2016deep}, MobileNet~\cite{howard2017mobilenets}, and so on.

\subsection{Inference with the Trained CNN Model}
During the inference phase, the trained CNN model is used to recognize the next character iteratively. In each iteration, the updated SuperOCR image is fed into the same CNN model. The entire characters reading is the sequential combination of all the recognized characters.

For OCR applications on edge devices, power consumption is strictly constrained. One advantage of the proposed SuperOCR method is being able to be deployed on a low-power CNN Domain-Specific Accelerator (CNN-DSA) chip~\cite{sun2018ultra,sun2018mram} connected to a low-end CPU, such as the Raspberry Pi (RBP). The CNN-DSA works as a coprocessor, which takes the heavy-lifting computation for the CNN inference. The RBP works as the host to generate the SuperOCR images, sending the image to the CNN-DSA, receiving the prediction for the next character, and returning the the sequencial combination of the characters as the final OCR result.

\section{Experimental Results}
The proposed SuperOCR method are applied to two application scenarios. One is the license plate recognition, the other is the watermeter characters recognition. Both applications have the fixed number of characters to be recognized, and the locations of the characters are random in each image.

\subsection{License Plate Recognition}
In this experiment, CCPD~\cite{xu2018towards} dataset is used. CCPD is a large and comprehensive dataset for car license plate recognition. It has more than 250k unique car images taken and mannually labeled by more than 800 Parking Fee Collector (PFC) from a roadside parking company. The images are taken by a handheld POS (Point-of-Sale) machine, so the location of the characters are random distributed in the image. In each image, the car license plate contains seven characters, including a Chinese character, a letter, and five numbers/letters. Images of uncontrolled conditions are also included to test the robustness of OCR models, such as rotation, distortion, illumination, snow or fog weather, and so on. 

\subsubsection{Experimental Setup} 
\label{CCPDsetup}

The 250k unique images of CCPD data are split into train, validation, and test as done in~\cite{xu2018towards}. The total 200k images of the CCPD-Base subset are split into Base-Train (100k) and Base-Val (100k). The paper~\cite{xu2018towards} mentions that the CCPD-Base is divided into two equal parts, but it does not describe the details of how to divide. The corresponding code in its github \footnote {https://github.com/detectRecog/CCPD} also does not release the details. So we randomly split CCPD-Base into two halves as training and validation respectively. The other subsets of CCPD are also used for evaluation as done in the CCPD paper, including CCPD-DB(Dark or Bright illuminations, 20k), CCPD-FN(Far or Near distance, 20k), CCPD-Rotate(Rotated, 10k), CCPD-Tilt(Tilted, 10k), CCPD-Weather(Image taken in a rainy, snowy or foggy day, 10k), and CCPD-Challenge(Challenging OCR images, 10k). 

There are totally 65 unique characters in the license plates of all the images. The PolyNet~\cite{zhang2017polynet} model pretrained on ImageNet is used to finetune on these 65 classes in this experiment. The pretrained PolyNet model has a relatively larger input resolution of 331x331 than the majority of the pretrained models of 224x224. However, it is still smaller than the input resolution of 480x480 for models used in ~\cite{xu2018towards}. The 331x331 input image has less than 70\% in length for each side of the 480x480 image, and less than 50\% of the area of pixels.

The SuperOCR image in this experiment is designed in two steps. First, the original input image is resized to occupy the upper potion of the SuperOCR image. In this experiment, the size of 331x305 is designed for the resized image, in order to best reserve the original image information, and at the same time leave abundent space for embedding the predicted characters. Second, only the first six characters on the license plate are embedded on the SuperOCR image, in order for a larger font size. The size of the generated SuperOCR dataset is seven times as big as the original dataset in CCPD.

A single 1080Ti (11GB memory, 250W) GPU is used, with batch size of 5 for training. Base learning rate is set as 1e-4, with a dropping rate of 0.99 every 5000 iterations. The weight decay is 7e-4 and momentum is 0.9. No data augumentation is used for training the SuperOCR model. On the contary, RPnet~\cite{xu2018towards} augments the training data by randomly sampling four times on each image to increase the training set by five times, which is helpful to training. Thus the future work could include data augumentation and larger memory and multiple GPUs to further boost the accuracy of SuperOCR model. 

\subsubsection{Recognition Result}
Table~\ref{CCPDcomparison} shows the OCR accuracy of the proposed SuperOCR method on the subsets of the CCPD dataset, including validation dataset (Base 100k), Rotate, Tilt, Weather, DB (Dark Bright), FN (Far Near), and Challenge. The recognition result of existing methods are also listed, including SSD300~\cite{liu2016ssd}+HC, YOLO9000~\cite{redmon2017yolo9000}+HC, FasterRCNN~\cite{ren2015faster}+HC, TE2E~\cite{li2018toward}, and RPnet~\cite{xu2018towards}. For a fair comparison, the resolution of the model input image is also listed in the second column. The resolution with the smallest input resolution is highlighted in bold font. For the other columns, the highest accuracy is highlighted. The result of other existing methods are borrowed from \cite{xu2018towards}. Note that \cite{xu2018towards} defines that the recoginition is correct if and only if IoU of the license plate detection is greater than 0.6 and all characters in the license plate are correctly recognized. On the contrary, since there is no detection module in the proposed SuperOCR method, the correctness of a recognition does not depend on the IoU of the detection result. When calculating the recognition accuracy of the proposed SuperOCR method, a recognition is defined as correct if all characters in the license plate are correctly recognized.

From the table we can see that the proposed SuperOCR method obtains slightly better result of 98.7\% on Base(100k) evaluation dataset than the previous state-of-the-art of 98.5\%. The Base dataset is relatively easy to recognize since the images are taken under controled scenarios. However, under some uncontroled scenarios, the recognition results given by the proposed SuperOCR method are significantly much better than the existing methods. For example, for images with rotated license plate, the accuracy of the proposed SuperOCR method is 95.6\%, which is 4.8\% more accurate than the best existing method. For tilted images of license plate, the accuracy is 96.0\%, which is 3.5\% better than the best existing method. For images taken in snowy or foggy weather, 6.8\% accuracy improvement is observed by using the SuperOCR method, i.e. improving the accuracy from the previous state-of-the-art 87.9\% to 94.7\%. For the most challenging images, as well as the dark and bright images, the proposed SuperOCR still obtains better accuracy than FasterRCNN+HC (Holistic-CNN~\cite{vspavnhel2017holistic}). The SuperOCR method performs worst on the far and near images, which is mainly caused by the smaller input resolution of 331x331 and the even smaller original image on the SuperOCR image. In the SuperOCR image, the original image only occupies part of the whole image of 331x331, thus the license plate is shrinked too much compared with the 480x480 input image. As a result, the license plate characters are even more difficult to recognize, especially for the images taken in a far distance. Alternating to a model with larger input size will help to improve the performance of recognition on images taken in a far distance.

\begin{figure*}[htb]
  \centering
  \centerline{\includegraphics[width=17.5cm]{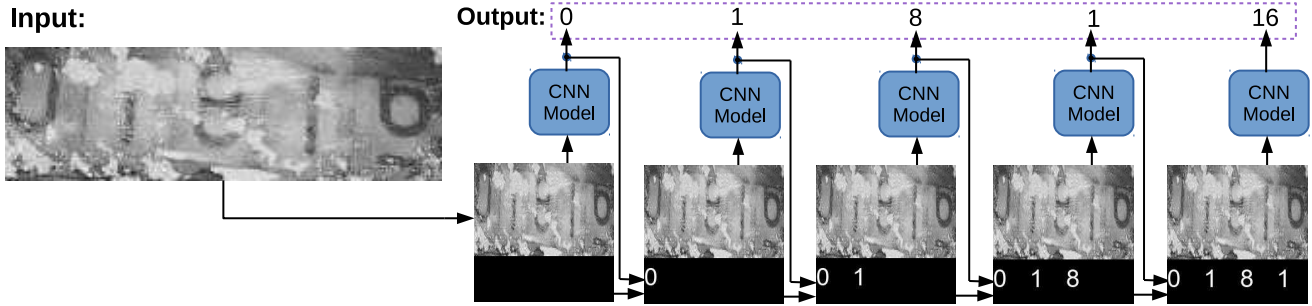}}
\caption{An example of SuperOCR method for watermeter number recognition. The input image may be tilted and/or blured. There are 20 possible classes for each digit, including the regular ten classes for digits from 0 through 9, and also the intermediate ten classes. For example, the state between 0 and 1 is the 10th class with a reading of ``0.5", and the state between 6 and 7 is the 16th class with a reading of ``6.5". The final recognition result of this example is ``01816.5".}
\label{figWatermeter}
\end{figure*}

\subsection{Watermeter Characters Recognition}
In this experiment, the Watetmeter Number Recognition dataset (WNR) in~\cite{yang2019fully} is used, which is downloadable at its corresponding github \footnote{https://github.com/jiarenyf/Water-Meter-Number-DataSet}. The WNR dataset is released by Deep Leaning and Visual Computing Lab of South China University of Technology for the research of watermeter number recognition. It consists of two parts. The first part contains 1000 clean images, and the second part contains 5000 noisy images with variant illumination, refraction, and occulation. The sizes of these images are different, with the width in range [201, 418], and height in range [37, 111]. This results in a relatively large range of aspective ratio in [2.933, 6.619]. In each image, there are five characters representing the reading from a mechanical watermeter. Besides the regular 10 classes for digits of 0 through 9, the ``mid-states" which representing the states in-between two consecutive digits are also counted as another 10 classes. Thus each character can be one of the 20 classes. 

\subsubsection{Experimental Setup} 
Following the convention in~\cite{yang2019fully}, a label $C$ in the range of [10, 19] denotes a intermediate state, while the range of [0, 9] denotes a regular state. For example, $C$=10 denotes the intermediate state between 0 and 1, while $C$=11 denotes the intermediate state between 1 and 2, and so on. If one of the character is an intermediate state, the final watermeter reading of the entire five characters will depend on the position of this intermediate state character. Specifictly, if this intermediate state character is at the end of the watermeter reading, the final reading will be interpreted as $C-9.5$. Otherwise, it will be interpreted as $C-10$.

As done in~\cite{yang2019fully}, the same 1000 of the 5000 noisy images are used for testing, and the remaining 4000 of the 5000 noisy images, plus all the 1000 clean images are used for training. In our experiment we only use the original images without any data augumentation.

Fig.~\ref{figWatermeter} shows the SuperOCR method for this WNR dataset. The size of SuperOCR image used for this experiment is 224x224. In each image, the top 224x150 pixels are designed to hold the resized original WNR image, and the bottom of the SuperOCR image is reserved for the first 4 characters. Similar to the above license plate recognition experiment, the last character will save the pixel area for the other characters to have a larger font size. Thus the last character is not be printed on the SuperOCR image. 

\subsubsection{Implementation on Edge Devices}
For IoT (Internet of Things) applications like watermeter characters recognition, the desired hardware implementations are edge devices with low power and low cost. We implement the SuperOCR method on a CNN-DSA chip~\cite{sun2018ultra} connected to a Raspberry Pi.  The Raspberry Pi, which can also be replaced by other low-end MCUs, works as a host to generate the SuperOCR image and send it to the CNN-DSA chip. And the CNN-DSA chip works as the coprocessor to execute the CNN model. A quantized model Gnet18~\cite{DBLP:conf/kdd/ShaSYZY19} is used in this exeriment. Gnet18 is a modified ResNet18 model with all shortcut removed in order to fit into the CNN accelerator~\cite{sun2018ultra,sun2018mram}. After the CNN layers, the FC layers are implemented on the host for the output prediction. The heavy-lifting work of CNN computations are done on the CNN-DSA chip, and the computational power for the FC layers required by the host CPU is negligible.

\subsubsection{Recognition Result}
\begin{table}
\begin{center}
\begin{tabular}{|c|c|c|}
\hline
Method   & LCR\% & AR\% \\
\hline
ResNet18 & 81.70 & 95.92 \\
CharNet & 84.10 & 96.64 \\
CRNN~\cite{shi2016end} & 86.10 & 97.04 \\
ConvS2S\cite{gehring2017convolutional} & 85.80 & 97.28 \\
FCSRN\cite{yang2019fully} & 85.60 & 96.98 \\
SRN\cite{yang2019fully}   & 89.60 & 97.82    \\
\hline
SuperOCR(CNN-DSA) & {\bf 90.50} & {\bf 97.98}\% \\

\hline
\end{tabular}
\end{center}
\caption{Accuracy comparison on watermeter dataset. The results of the first six methods are given by~\cite{yang2019fully}, where AR is defined as the Accuracy Rate, and LCR is defined as the Line Correct Rate to denote the accuracy for the whole line of characters. The SuperOCR implementation on the CNN-DSA implementation uses the CNN accelerator chip with 300mW power consumption.}
\label{WatermeterComparison}
\end{table}

The recognition result on WNR dataset is shown in Table~\ref{WatermeterComparison}. The first six rows of result are given by~\cite{yang2019fully}, and the last row shows the result given by the SuperOCR method implemented on the low-power CNN-DSA chip. It can be seen that the proposed SuperOCR overperforms all the other existing methods on both LCR and AR metrics. LCR (Line Correct Rate) and AR (Accuracy Rate) are defined in~\cite{yin2013icdar}. The LCR calculate the accuracy at the image level, i.e. an image will be counted as correct if and only if all the characters in this image is correct. On the contrary, AR calculate the accuracy at the character level.

For realworld applications, accuracy is not the only criteria for the system performance. Power and cost are also the very important factors when comparing different solutions. The proposed SuperOCR method is implemented on the low-power and low-cost edge-device to show its feasibility as a practical method in a real system. The other existing methods in Table~\ref{WatermeterComparison} are not able to be directly deployed into a CNN-DSA chip. If they are deployed on a 1080Ti GPU, the power consumption for inference is around 90W with a maximum of 2.5GB memory footprint. As a comparison, the proposed SuperOCR method implemented on the CNN-DSA chip only consumes 300mW power, which is 300x power saving than the GPU device. The on-chip memory of the CNN-DSA is only 9MB. 

\section{Conclusion}
In this paper, we propose the SuperOCR method to recognize the characters without any detection module. This is done by converting the OCR task into an image captioning task, and borrows the concept of two-dimensional embedding from the Super Characters method applied in Natural Language Processing tasks. This is the first work of casting the OCR problem into an image captioning problem. The proposed method could be fine-tuned from a CNN model pretrained on large image recognition dataset. Experimental results shows the proposed method outperforms other existing methods. In addition, we also deploy the method on Raspberry Pi 3 with a low-power CNN accelerator for edge applications.


{\small
\bibliographystyle{ieee_fullname}
\bibliography{egbib}

\begin{thebibliography}{10}\itemsep=-1pt

\bibitem{anderson2018bottom}
Peter Anderson, Xiaodong He, Chris Buehler, Damien Teney, Mark Johnson, Stephen
  Gould, and Lei Zhang.
\newblock Bottom-up and top-down attention for image captioning and visual
  question answering.
\newblock In {\em Proceedings of the IEEE conference on computer vision and
  pattern recognition}, pages 6077--6086, 2018.

\bibitem{deng2009imagenet}
Jia Deng, Wei Dong, Richard Socher, Li-Jia Li, Kai Li, and Li Fei-Fei.
\newblock Imagenet: A large-scale hierarchical image database.
\newblock In {\em 2009 IEEE conference on computer vision and pattern
  recognition}, pages 248--255. Ieee, 2009.

\bibitem{gehring2017convolutional}
Jonas Gehring, Michael Auli, David Grangier, Denis Yarats, and Yann~N Dauphin.
\newblock Convolutional sequence to sequence learning.
\newblock In {\em Proceedings of the 34th International Conference on Machine
  Learning-Volume 70}, pages 1243--1252. JMLR. org, 2017.

\bibitem{he2016deep}
Kaiming He, Xiangyu Zhang, Shaoqing Ren, and Jian Sun.
\newblock Deep residual learning for image recognition.
\newblock In {\em Proceedings of the IEEE conference on computer vision and
  pattern recognition}, pages 770--778, 2016.

\bibitem{hossain2019comprehensive}
MD~Zakir Hossain, Ferdous Sohel, Mohd~Fairuz Shiratuddin, and Hamid Laga.
\newblock A comprehensive survey of deep learning for image captioning.
\newblock {\em ACM Computing Surveys (CSUR)}, 51(6):1--36, 2019.

\bibitem{howard2017mobilenets}
Andrew~G Howard, Menglong Zhu, Bo Chen, Dmitry Kalenichenko, Weijun Wang,
  Tobias Weyand, Marco Andreetto, and Hartwig Adam.
\newblock Mobilenets: Efficient convolutional neural networks for mobile vision
  applications.
\newblock {\em arXiv preprint arXiv:1704.04861}, 2017.

\bibitem{johnson2016densecap}
Justin Johnson, Andrej Karpathy, and Li Fei-Fei.
\newblock Densecap: Fully convolutional localization networks for dense
  captioning.
\newblock In {\em Proceedings of the IEEE Conference on Computer Vision and
  Pattern Recognition}, pages 4565--4574, 2016.

\bibitem{karpathy2015deep}
Andrej Karpathy and Li Fei-Fei.
\newblock Deep visual-semantic alignments for generating image descriptions.
\newblock In {\em Proceedings of the IEEE conference on computer vision and
  pattern recognition}, pages 3128--3137, 2015.

\bibitem{laroca2018robust}
Rayson Laroca, Evair Severo, Luiz~A Zanlorensi, Luiz~S Oliveira,
  Gabriel~Resende Gon{\c{c}}alves, William~Robson Schwartz, and David Menotti.
\newblock A robust real-time automatic license plate recognition based on the
  yolo detector.
\newblock In {\em 2018 International Joint Conference on Neural Networks
  (IJCNN)}, pages 1--10. IEEE, 2018.

\bibitem{laroca2019efficient}
Rayson Laroca, Luiz~A Zanlorensi, Gabriel~R Gon{\c{c}}alves, Eduardo Todt,
  William~Robson Schwartz, and David Menotti.
\newblock An efficient and layout-independent automatic license plate
  recognition system based on the yolo detector.
\newblock {\em arXiv preprint arXiv:1909.01754}, 2019.

\bibitem{lecun1998mnist}
Yann LeCun.
\newblock The mnist database of handwritten digits.
\newblock {\em http://yann. lecun. com/exdb/mnist/}, 1998.

\bibitem{li2018toward}
Hui Li, Peng Wang, and Chunhua Shen.
\newblock Toward end-to-end car license plate detection and recognition with
  deep neural networks.
\newblock {\em IEEE Transactions on Intelligent Transportation Systems},
  20(3):1126--1136, 2018.

\bibitem{liu2011casia}
Cheng-Lin Liu, Fei Yin, Da-Han Wang, and Qiu-Feng Wang.
\newblock Casia online and offline chinese handwriting databases.
\newblock In {\em 2011 International Conference on Document Analysis and
  Recognition}, pages 37--41. IEEE, 2011.

\bibitem{liu2016ssd}
Wei Liu, Dragomir Anguelov, Dumitru Erhan, Christian Szegedy, Scott Reed,
  Cheng-Yang Fu, and Alexander~C Berg.
\newblock Ssd: Single shot multibox detector.
\newblock In {\em European conference on computer vision}, pages 21--37.
  Springer, 2016.

\bibitem{lu2017knowing}
Jiasen Lu, Caiming Xiong, Devi Parikh, and Richard Socher.
\newblock Knowing when to look: Adaptive attention via a visual sentinel for
  image captioning.
\newblock In {\em Proceedings of the IEEE conference on computer vision and
  pattern recognition}, pages 375--383, 2017.

\bibitem{masood2017license}
Syed~Zain Masood, Guang Shu, Afshin Dehghan, and Enrique~G Ortiz.
\newblock License plate detection and recognition using deeply learned
  convolutional neural networks.
\newblock {\em arXiv preprint arXiv:1703.07330}, 2017.

\bibitem{netzer2011reading}
Yuval Netzer, Tao Wang, Adam Coates, Alessandro Bissacco, Bo Wu, and Andrew~Y
  Ng.
\newblock Reading digits in natural images with unsupervised feature learning.
\newblock 2011.

\bibitem{redmon2017yolo9000}
Joseph Redmon and Ali Farhadi.
\newblock Yolo9000: better, faster, stronger.
\newblock In {\em Proceedings of the IEEE conference on computer vision and
  pattern recognition}, pages 7263--7271, 2017.

\bibitem{ren2015faster}
Shaoqing Ren, Kaiming He, Ross Girshick, and Jian Sun.
\newblock Faster r-cnn: Towards real-time object detection with region proposal
  networks.
\newblock In {\em Advances in neural information processing systems}, pages
  91--99, 2015.

\bibitem{rennie2017self}
Steven~J Rennie, Etienne Marcheret, Youssef Mroueh, Jerret Ross, and Vaibhava
  Goel.
\newblock Self-critical sequence training for image captioning.
\newblock In {\em Proceedings of the IEEE Conference on Computer Vision and
  Pattern Recognition}, pages 7008--7024, 2017.

\bibitem{rizvi2017deep}
Syed Tahir~Hussain Rizvi, Denis Patti, Tomas Bj{\"o}rklund, Gianpiero Cabodi,
  and Gianluca Francini.
\newblock Deep classifiers-based license plate detection, localization and
  recognition on gpu-powered mobile platform.
\newblock {\em Future Internet}, 9(4):66, 2017.

\bibitem{selmi2017deep}
Zied Selmi, Mohamed~Ben Halima, and Adel~M Alimi.
\newblock Deep learning system for automatic license plate detection and
  recognition.
\newblock In {\em 2017 14th IAPR International Conference on Document Analysis
  and Recognition (ICDAR)}, volume~1, pages 1132--1138. IEEE, 2017.

\bibitem{DBLP:conf/kdd/ShaSYZY19}
Hao Sha, Baohua Sun, Nicholas Yi, Wenhan Zhang, and Lin Yang.
\newblock On-device chatbot system using superchat method on raspberry pi and
  {CNN} domain specific accelerator.
\newblock In Albert Bifet, Michele Berlingerio, Jo{\~{a}}o Gama, Jesse Read,
  and Ana~Rita Nogueira, editors, {\em Proceedings of the 8th International
  Workshop on Big Data, IoT Streams and Heterogeneous Source Mining:
  Algorithms, Systems, Programming Models and Applications co-located with 25th
  {ACM} {SIGKDD} Conference on Knowledge Discovery and Data Mining {(KDD}
  2019), Anchorage, Alaska, August 4-8, 2019}, volume 2579 of {\em {CEUR}
  Workshop Proceedings}. CEUR-WS.org, 2019.

\bibitem{shi2016end}
Baoguang Shi, Xiang Bai, and Cong Yao.
\newblock An end-to-end trainable neural network for image-based sequence
  recognition and its application to scene text recognition.
\newblock {\em IEEE transactions on pattern analysis and machine intelligence},
  39(11):2298--2304, 2016.

\bibitem{silva2017real}
Sergio~Montazzolli Silva and Claudio~Rosito Jung.
\newblock Real-time brazilian license plate detection and recognition using
  deep convolutional neural networks.
\newblock In {\em 2017 30th SIBGRAPI Conference on Graphics, Patterns and
  Images (SIBGRAPI)}, pages 55--62. IEEE, 2017.

\bibitem{simonyan2014very}
Karen Simonyan and Andrew Zisserman.
\newblock Very deep convolutional networks for large-scale image recognition.
\newblock {\em arXiv preprint arXiv:1409.1556}, 2014.

\bibitem{vspavnhel2017holistic}
Jakub {\v{S}}pa{\v{n}}hel, Jakub Sochor, Roman Jur{\'a}nek, Adam Herout,
  Luk{\'a}{\v{s}} Mar{\v{s}}{\'\i}k, and Pavel Zem{\v{c}}{\'\i}k.
\newblock Holistic recognition of low quality license plates by cnn using track
  annotated data.
\newblock In {\em 2017 14th IEEE International Conference on Advanced Video and
  Signal Based Surveillance (AVSS)}, pages 1--6. IEEE, 2017.

\bibitem{sun2018mram}
Baohua Sun, Daniel Liu, Leo Yu, Jay Li, Helen Liu, Wenhan Zhang, and Terry
  Torng.
\newblock Mram co-designed processing-in-memory cnn accelerator for mobile and
  iot applications.
\newblock {\em arXiv preprint arXiv:1811.12179}, 2018.

\bibitem{sun2018super}
Baohua Sun, Lin Yang, Patrick Dong, Wenhan Zhang, Jason Dong, and Charles
  Young.
\newblock Super characters: A conversion from sentiment classification to image
  classification.
\newblock In {\em Proceedings of the 9th Workshop on Computational Approaches
  to Subjectivity, Sentiment and Social Media Analysis}, pages 309--315, 2018.

\bibitem{sun2018ultra}
Baohua Sun, Lin Yang, Patrick Dong, Wenhan Zhang, Jason Dong, and Charles
  Young.
\newblock Ultra power-efficient cnn domain specific accelerator with 9.3
  tops/watt for mobile and embedded applications.
\newblock In {\em Proceedings of the IEEE Conference on Computer Vision and
  Pattern Recognition Workshops}, pages 1677--1685, 2018.

\bibitem{sun2019supercaptioning}
Baohua Sun, Lin Yang, Michael Lin, Charles Young, Patrick Dong, Wenhan Zhang,
  and Jason Dong.
\newblock Supercaptioning: Image captioning using two-dimensional word
  embedding.
\newblock {\em arXiv preprint arXiv:1905.10515}, 2019.

\bibitem{tan2019efficientnet}
Mingxing Tan and Quoc~V Le.
\newblock Efficientnet: Rethinking model scaling for convolutional neural
  networks.
\newblock {\em arXiv preprint arXiv:1905.11946}, 2019.

\bibitem{vinyals2016show}
Oriol Vinyals, Alexander Toshev, Samy Bengio, and Dumitru Erhan.
\newblock Show and tell: Lessons learned from the 2015 mscoco image captioning
  challenge.
\newblock {\em IEEE transactions on pattern analysis and machine intelligence},
  39(4):652--663, 2016.

\bibitem{xu2018towards}
Zhenbo Xu, Wei Yang, Ajin Meng, Nanxue Lu, Huan Huang, Changchun Ying, and
  Liusheng Huang.
\newblock Towards end-to-end license plate detection and recognition: A large
  dataset and baseline.
\newblock In {\em Proceedings of the European Conference on Computer Vision
  (ECCV)}, pages 255--271, 2018.

\bibitem{yang2019fully}
Fan Yang, Lianwen Jin, Songxuan Lai, Xue Gao, and Zhaohai Li.
\newblock Fully convolutional sequence recognition network for water meter
  number reading.
\newblock {\em IEEE Access}, 7:11679--11687, 2019.

\bibitem{yao2017boosting}
Ting Yao, Yingwei Pan, Yehao Li, Zhaofan Qiu, and Tao Mei.
\newblock Boosting image captioning with attributes.
\newblock In {\em Proceedings of the IEEE International Conference on Computer
  Vision}, pages 4894--4902, 2017.

\bibitem{yin2013icdar}
Fei Yin, Qiu-Feng Wang, Xu-Yao Zhang, and Cheng-Lin Liu.
\newblock Icdar 2013 chinese handwriting recognition competition.
\newblock In {\em 2013 12th International Conference on Document Analysis and
  Recognition}, pages 1464--1470. IEEE, 2013.

\bibitem{you2016image}
Quanzeng You, Hailin Jin, Zhaowen Wang, Chen Fang, and Jiebo Luo.
\newblock Image captioning with semantic attention.
\newblock In {\em Proceedings of the IEEE conference on computer vision and
  pattern recognition}, pages 4651--4659, 2016.

\bibitem{zhang2017polynet}
Xingcheng Zhang, Zhizhong Li, Chen Change~Loy, and Dahua Lin.
\newblock Polynet: A pursuit of structural diversity in very deep networks.
\newblock In {\em Proceedings of the IEEE Conference on Computer Vision and
  Pattern Recognition}, pages 718--726, 2017.

\end{thebibliography}
}

\end{document}